\documentclass[10pt,journal,cspaper,compsoc]{IEEEtran}
\usepackage{amssymb}

\ifCLASSOPTIONcompsoc
  \usepackage{cite}
\else
\fi
\ifCLASSINFOpdf
  \usepackage[pdftex]{graphicx}
\else
\fi
\usepackage{amsmath}
\newcommand{\argmin}{\operatornamewithlimits{\arg min}}

\usepackage{algpseudocode}

\newsavebox{\ieeealgbox}

\usepackage{array}

\usepackage{mdwmath}
\usepackage{mdwtab}

\usepackage{eqparbox}

\usepackage{caption}
\usepackage{subcaption}

\usepackage{fixltx2e}

\usepackage{stfloats}

\ifCLASSOPTIONcaptionsoff
 \usepackage[nomarkers]{endfloat}
\let\MYoriglatexcaption\caption
\renewcommand{\caption}[2][\relax]{\MYoriglatexcaption[#2]{#2}}
\fi
\usepackage{url}

\newcommand{\qed}{\nobreak \ifvmode \relax \else
  \ifdim\lastskip<1.5em \hskip-\lastskip
  \hskip1.5em plus0em minus0.5em \fi \nobreak
  \vrule height0.75em width0.5em depth0.25em\fi}

\begin{document}
%
\title{Robust Text Detection in Natural Scene Images}
%
%
%
%

\author{Xu-Cheng Yin,
       Xuwang Yin,
       Kaizhu Huang,
       and Hong-Wei Hao

\IEEEcompsocitemizethanks{\IEEEcompsocthanksitem X.-C. Yin is with the Department
of Computer Science and Technology, School of Computer and Communication Engineering, University of Science and Technology Beijing,
Beijing 100083, China.\protect\\
E-mail: xuchengyin@ustb.edu.cn.
\IEEEcompsocthanksitem X. Yin is with the Department
of Computer Science and Technology,
School of Computer and Communication Engineering, University of Science and Technology Beijing, Beijing 100083, China.
\IEEEcompsocthanksitem K. Huang is with Department of Electrical and Electronic Engineering, Xi'an Jiaotong-Liverpool University, Suzhou 215123, China.
\IEEEcompsocthanksitem H.-W. Hao is with Institute of Automation, Chinese Academy of Sciences, Beijing 100190, China.
}
\thanks{}}

\IEEEcompsoctitleabstractindextext{%
  \begin{abstract}
    Text detection in natural scene images is an
important prerequisite for many content-based image
analysis tasks. In this paper, we propose an accurate and robust method for detecting texts
in natural scene images.
    A fast and effective pruning algorithm is designed to
    extract Maximally Stable Extremal Regions (MSERs) as
    character candidates using the strategy of minimizing
    regularized variations.
    Character candidates are grouped into text candidates
    by the single-link clustering algorithm, where distance weights and
    threshold of the clustering algorithm are learned automatically by a novel self-training distance metric learning algorithm.
    The posterior probabilities of text candidates corresponding
    to non-text are estimated with an character classifier;
    text candidates with high probabilities are then eliminated and finally
    texts are identified with a text classifier.
    The proposed system is evaluated on the ICDAR 2011 Robust
    Reading Competition dataset; the $f$ measure is over
    76\% and is significantly better than the state-of-the-art performance
    of 71\%. Experimental results on a publicly available multilingual dataset
    also show that our proposed method can outperform the other competitive method with the  $f$ measure  increase of over $9$ percent.
    Finally, we have setup an online demo of our proposed scene text detection
    system at \emph{``http://kems.ustb.edu.cn/learning/yin/dtext"}.

\end{abstract}

}

\maketitle

\IEEEdisplaynotcompsoctitleabstractindextext

%
\IEEEpeerreviewmaketitle

\section{Introduction}

\IEEEPARstart{T}{ext} in images contains
valuable information and is exploited in many
content-based image and video applications, such as content-based web
image search, video information retrieval, mobile based text analysis and
recognition~\cite{Zhong2000, Doermann2000, Weinman2009, Yin2011, Chew2011}.
Due to complex background, variations of  font,
size, color and orientation, text in natural scene images
has to be robustly detected before being recognized or retrieved.

Existing methods for scene text detection can  roughly be
categorized into three groups: sliding window based
methods~\cite{derivative_feature, adaboost_text, Kim2003},
connected component based methods~\cite{Epshtein, structure-partition, color-clustering},
and hybrid methods~\cite{pan}.
Sliding window based methods, also called as region based methods, engage a sliding window to search
for possible texts in the image and then use machine learning
techniques to identify texts.
These methods tend to be slow as the image has to be
processed in multiple scales.
Connected component based methods extract character candidates from
images by connected component analysis followed by grouping
character candidates into text; additional checks may be
performed to remove false positives.
The hybrid method presented by Pan et al.~\cite{pan} exploits a
region detector to detect text candidates and extracts
connected components as character candidates by local
binarization; non-characters are eliminated with a Conditional Random Fields~(CRFs)~\cite{crf} model,
 and characters can finally be grouped into text.
More recently, Maximally Stable Extremal Region (MSER) based methods, which actually fall into the family of
connected component based methods but use MSERs~\cite{mser} as character candidates, have
become the focus of several recent projects~\cite{icdar2011, edge_mser, head_mounted,
  real_time, pruned_search, neumann_method, mser2013}.

Although the MSER based method is the winning method of the benchmark data, i.e., ICDAR
2011 Robust Reading Competition~\cite{icdar2011}
and has reported promising
performance, there remains several problems to be addressed.
First, as the MSER algorithm detects a large number of non-characters,
most of the character candidates need to be removed
before further processing. The existing methods for MSERs
pruning~\cite{head_mounted, real_time},  on one hand, may still have room for further improvement in terms of the accuracies;  on the other hand, they tend to be slow because
of the computation of complex features.
Second, current approaches~\cite{head_mounted, real_time, pan} for
text candidates construction, which can be categorized as
rule based and clustering based methods, work well but
are still not sufficient;
rule based methods generally require hand-tuned parameters,
which is time consuming and error pruning;
the clustering based method~\cite{pan} shows good performance but it is
complicated by incorporating a second stage processing
after minimum spanning tree clustering.

In this paper, we propose a robust and accurate MSER based scene text detection
method.
First, by exploring the hierarchical structure of MSERs and
adopting simple features, we designed a fast and accurate MSERs
pruning algorithm;
the number of character candidates to be processed is
significantly reduced with a high accuracy.
Second, we propose a novel self-training distance metric learning algorithm
that can learn distance weights and clustering threshold simultaneously and automatically;
character candidates are clustered into text candidates
by the single-link clustering algorithm using the learned parameters.
Third, we propose to use a character classifier to
estimate the posterior probabilities of text candidates
corresponding to non-text and remove text candidates with
high probabilities.
Such elimination helps to train a more powerful text
classifier for identifying text.
By integrating the above ideas, we built an accurate and robust
scene text detection system.
The system is evaluated on the benchmark ICDAR 2011 Robust Reading
Competition dataset and achieved an $f$ measure of 76\%. To our best knowledge, this result ranks the first on all the reported performance and is much higher the current best performance of 71\%.
We also validate our method on the multilingual (include Chinese and English) dataset used in ~\cite{pan}. With an $f$ measure of 74.58\%, our system  significantly outperforms  the competitive method~\cite{pan} that achieves only 65.2\%.
An online demo of our proposed scene text detection system
  is available at \emph{\url{http://kems.ustb.edu.cn/learning/yin/dtext}}.

The rest of this paper is organized as follows. Recent MSER
based scene text detection methods
are reviewed in Section~\ref{section:related_work}.
Section~\ref{section:system} describes the proposed scene text detection method.
Section~\ref{section:experimental_results} presents the experimental results of the proposed system
on ICDAR 2011 Robust Reading Competition dataset and a
multilingual (include Chinese and English) dataset.
Final remarks are presented in Section~\ref{section:conclusion}.

\section{Related Work}
\label{section:related_work}
As described above,  MSER based methods have demonstrated very promising performance in many real projects. However, current MSER based methods still have some
key limitations, i.e., they may suffer from  large number of non-characters candidates in detection and also
insufficient text candidates construction algorithms. In this section, we review the MSER
based methods with the focus on these two problems. Other scene text detection methods can be referred to in some
survey papers~\cite{survey04,survey05,survey08}. A recently published MSER based method can be referred to in
 Shi et al.~\cite{mser2013}.

The main advantage of MSER based methods over traditional
connected component based methods may root in the usage of MSERs as
character candidates.
Although the MSER algorithm can detect most characters even when
the image is in low quality (low resolution, strong noises,
low contrast, etc.), most of the detected character candidates correspond
to non-characters.
Carlos et al.~\cite{head_mounted} presented a MSERs pruning
algorithm that contains two steps: (1) reduction of linear
segments and (2) hierarchical filtering.
The first stage reduces linear segments in the MSER tree into
one node by maximizing the \emph{border energy} function;
the second stage walks through the tree in a depth-first
manner and eliminates nodes by checking them against a
cascade of filters: \emph{size, aspect ratio, complexity,
  border energy and texture}.
Neumann and Matas~\cite{real_time} presented a two stage algorithm for
Extremal Regions (ERs) pruning.
In the first stage, a classifier trained from
incrementally computable descriptors (\emph{area, bounding
  box, perimeter, Euler number and horizontal crossing}) is used to
estimate the class-conditional probabilities
$p(r|\mbox{chracter})$ of ERs; ERs corresponding to
local maximum of probabilities in the
ER inclusion relation are selected.
In the second stage, ERs passed the first stage are
classified as characters and non-characters using more
complex features. As most of the MSERs correspond to
non-characters, the purpose of using cascading filters and
incrementally computable descriptors in these above two methods is
to deal with the computational complexity caused by the high
false positive rate.

Another challenge of MSER based methods, or more generally,
CC-based methods and hybrid methods, is how to group
character candidates into text candidates.
The existing methods for text candidates construction fall
into two general approaches: rule-based~\cite{edge_mser,
  head_mounted, real_time} and clustering-based methods~\cite{pan}.
Neumann and Matas~\cite{real_time} grouped character
candidates using the text line constrains, whose basic
assumption is that characters in a word can be fitted by one or
more top and bottom lines.
Carlos et al.~\cite{head_mounted} constructed a fully connected
graph over character candidates; they filtered edges by
running a set of tests (edge angle, relative position
and size difference of adjacent character candidates) and
used the remaining connected subgraphs as text candidates.
Chen et al.~\cite{edge_mser} pairwised character candidates as
clusters by putting constrains on stroke width and height
difference; they then
exploited a straight line to fit to the centroids of
clusters and declared a line as text candidate if it
connected three or more  character candidates.
The clustering-based method presented by Pan et al.~\cite{pan} clusters character
candidates into a tree using the minimum spanning tree
algorithm with a learned distance metric~\cite{yin-liu:metric-2009};
text candidates are constructed by cutting off between-text
edges with an energy minimization model.
The above rule-based methods generally require hand-tuned
parameters, while the clustering-based method
is complicated by the incorporating of the
post-processing stage, where one has to specify the
energy model.

\section{Robust Scene Text Detection}
\label{section:system}
\begin{figure}
  \centering
  \includegraphics[width=0.4\textwidth]{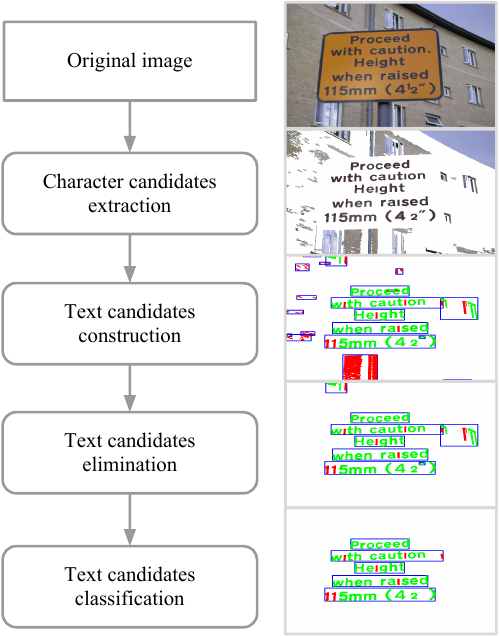}
  \caption{Flowchart of the proposed system and the corresponding experimental
    results after each step of a sample image.
    Text candidates are labeled by blue bounding rectangles;
    character candidates identified as characters are
    colored green, others red.
  }
  \label{fig:system_overview}
\end{figure}

In this paper, by incorporating several key improvements over
traditional MSER based methods, we propose a novel MSER based
scene text detection method, which finally leads to significant performance improvement over the other competitive methods.
The structure of the proposed system, as well as the sample result
of each stage is presented in
Figure~\ref{fig:system_overview}.
The proposed scene text detection method includes the
following stages:

1)  \emph{Character candidates extraction}.
character candidates are extracted using the MSER algorithm;
most of the non-characters are reduced by the proposed MSERs
pruning algorithm using the strategy of minimizing
regularized variations. More details are presented in Section~\ref{section:mser_extraction}.

2)  \emph{Text candidates construction}.
distance weights and threshold are learned simultaneously
using the proposed metric learning algorithm; character
candidates are clustered into text candidates by the
single-link clustering algorithm using the learned
parameters. More details are presented in Section~\ref{section:region_construction}.

3)  \emph{Text candidates elimination}.   the
posterior probabilities of text candidates corresponding to
non-text are measured using the character classifier and
text candidates with high probabilities for non-text are removed.
More details are presented in Section~\ref{section:character_classifier}.

4)  \emph{Text candidates classification}.   text
candidates corresponding to true text are identified by the
text classifier. An AdaBoost
classifier is trained to decide whether an text candidate corresponding
to true text or not~\cite{Yin12}.
As characters in the same text tend to have similar features,
the uniformity of character candidates'
features are used as text candidate's features to train the classifier.

In order to measure the performance of the proposed system
using the ICDAR 2011 competition dataset, text candidates
identified as text are further partitioned into words
by classifying inner character distances into character
spacings and word spacings using an AdaBoost classifier~\cite{Yin12}.
The following features are adopted: spacing aspect ratio,
relative width difference between left and right
neighbors, number of character candidates in the text
candidate.



\subsection{Letter Candidates Extraction}
\label{section:mser_extraction}
\subsubsection{Pruning Algorithm Overview}
The MSER algorithm is able to detect almost all
characters even when the image is in low quality.
However, as shown in Figure~\ref{fig:mser_tree_origin}, most of the detected
character candidates correspond to non-characters and
should be removed before further processing.
Figure~\ref{fig:mser_tree_origin} also shows that the detected characters forms
a tree, which is quite useful for designing  the
pruning algorithm.
 In real world situations, as characters cannot be
``included'' by or ``include''
other characters, it is safe to remove children once the
parent is known to be a character, and vice versa.
The parent-children elimination is a safe operation
because characters are preserved after the operation.
By reduction, if the MSER tree is pruned by applying
parent-children elimination operation
recursively in a depth-first manner, we are still in safe
place and characters are preserved.
As shown in Figure~\ref{fig:mser_tree_tree_accumulation},
the above algorithm will end up with a set of
disconnected nodes containing all the characters.
The problem with the above algorithm is that it is expensive to
identify character.
Fortunately, rather than identifying the character,
the choice between parent and children can be
made by simply choosing the one that is more likely to be
characters,
which can be estimated by
the proposed regularized variation scheme.
Considering different situations in MSER trees,
we design two versions of the parent-children
elimination method,
namely the \emph{linear reduction} and
\emph{tree accumulation} algorithm.
Non-character regions are eliminated by the linear
reduction and tree accumulation algorithm using the strategy
of minimizing regularized variations.
Our experiment on ICDAR 2011 competition training set shows that
more than 80\% of character candidates are eliminated using
the proposed pruning algorithm.

In the following sections, we first introduce the
concept of variation and explain why variations need to be
regularized. Then we introduce the linear reduction and
tree accumulation algorithm. Finally we present the
complexity analysis for the proposed algorithms.

\subsubsection{Variation and Its Regularization }
According to Matas et al.~\cite{mser},
an ``extremal region'' is a connected component of an image
whose pixels have either higher or lower intensity than its
outer boundary pixels~\cite{vlfeat, detector_compare}.
Extremal regions are extracted by applying a set of
increasing intensity levels to the gray scale image.
When the intensity level increases,
a new extremal region is extracted by
accumulating pixels of current level and joining
lower level extremal regions~\cite{head_mounted};
when the top level is reached, extremal regions of the whole image are extracted as
a rooted tree.
The variation of an extremal region is defined as follows.
Let $R_l$ be an extremal region, $B(R_l) = (R_l, R_{l+1}, ...,
R_{l+\Delta})$
($\Delta$ is an parameter)  be the branch of the tree rooted at $R_l$,
the variation (instability) of $R_l$ is defined as
\begin{equation}
  v(R_l) = \frac{|R_{l+\Delta} - R_l|}{|R_l|}.
\end{equation}
An extremal region $R_l$ is a maximally stable extremal region if its
variation is lower than (more stable) its parent $R_{l-1}$ and
child $R_{l+1}$~\cite{vlfeat, mser}.
Informally, a maximally stable extremal region is an extremal region
whose size remains virtually unchanged over a range of
intensity levels~\cite{real_time}.

\begin{figure}[htb!]
  \centering
  \begin{subfigure}[b]{0.11\textwidth}
    \centering
    \includegraphics[width=\textwidth]{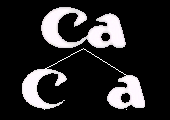}
    \caption{}
    \label{fig:situation1}
  \end{subfigure}
  \begin{subfigure}[b]{0.11\textwidth}
    \centering
    \includegraphics[width=\textwidth]{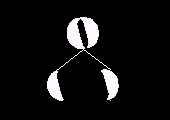}
    \caption{}
    \label{fig:situation2}
  \end{subfigure}
  \caption{Character correspondence in MSER trees.
    (a) A MSER tree whose children corresponds to characters;
    (b) a MSER tree whose parent corresponds to character.
  }
  \label{fig:situations}
\end{figure}

As MSERs with lower variations have sharper
borders
and are more likely to be characters,
one possible strategy may be used by the
parent-children elimination operation is to select parent or children
based on who have the lowest variation.
However, this strategy alone will not work because
MSERs corresponding to characters may not
necessarily have lowest variations.
Consider a very common situation depicted in Figure~\ref{fig:situations}.
The children of the MSER tree in Figure~\ref{fig:situation1} correspond to
characters while the parent of the MSRE tree in
Figure~\ref{fig:situation2} corresponds to
character.
The ``minimize variation'' strategy cannot deal with this
situation because either parent or children may
have the lowest variations.
However, our experiment shows that this limitation can be easily fixed by variation
regularization, whose  basic idea is to penalize variations of
MSERs with too large or too small aspect ratios.
Note that we are not requiring characters to have the lowest
variations globally, a lower variation in a parent-children
relationship suffices for our algorithm.

Let $\mathcal{V}$ be the variation and $a$ be the aspect
ratio of a MSER,
the aspect ratios of characters
are expected to fall in $[a_{min}, a_{max}]$, the
regularized variation is defined as
\begin{equation}
\mathcal{V} =
\begin{cases}
  \mathcal{V} - \theta_1 (a - a_{max}) & \text{if } a > a_{max} \\
  \mathcal{V} - \theta_2 (a_{min} - a) & \text{if } a < a_{min} \\
  \mathcal{V}                      & \text{otherwise}
\end{cases}\;,
\end{equation}
where $\theta_1$ and $\theta_2$ are penalty parameters.
Based on experiments on the training dataset,
these parameters are set as
$\theta_1 = 0.03, \theta_2 = 0.08, a_{max} = 1.2 \text{ and }
a_{min} = 0.7$.

\begin{figure}[htb!]
  \centering
  \begin{subfigure}[b]{0.09\textwidth}
    \centering
    \includegraphics[trim = 60mm 14mm 0mm 107mm, clip,
    width=\textwidth]{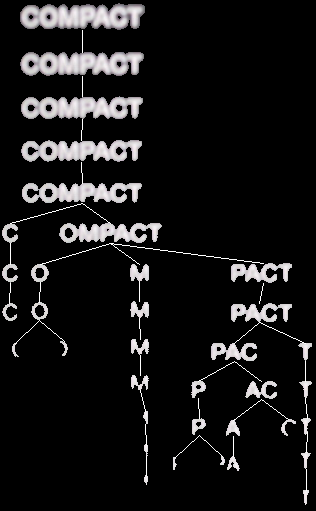}
    \caption{}
    \label{fig:mser_tree_origin}
  \end{subfigure}
  \begin{subfigure}[b]{0.09\textwidth}
    \centering
    \includegraphics[trim = 60mm 14mm 0mm 107mm, clip, width=\textwidth]{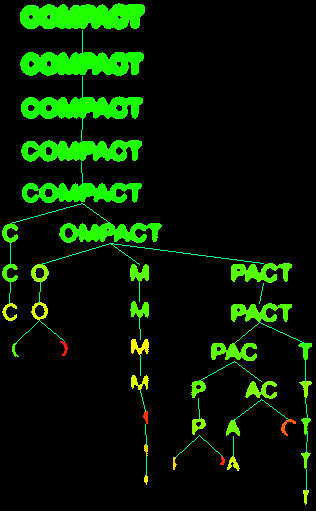}
    \caption{}
    \label{fig:mser_tree_var}
  \end{subfigure}
  \begin{subfigure}[b]{0.09\textwidth}
    \centering
    \includegraphics[trim = 60mm 14mm 0mm 107mm, clip, width=\textwidth]{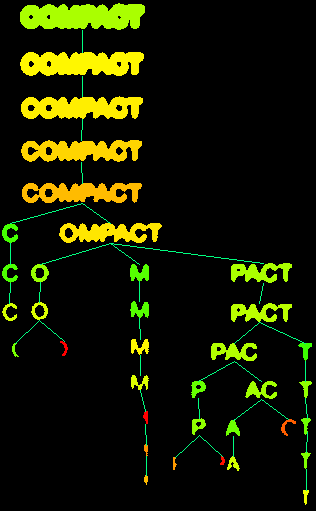}
    \caption{}
    \label{fig:mser_tree_regularized_var}
  \end{subfigure}
  \begin{subfigure}[b]{0.09\textwidth}
    \centering
    \includegraphics[trim = 60mm 0mm 0mm 35mm, clip, width=\textwidth]{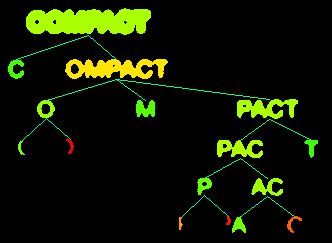}
    \caption{}
    \label{fig:mser_tree_linear_reduction}
  \end{subfigure}
  \begin{subfigure}[b]{0.09\textwidth}
    \centering
    \includegraphics[width=\textwidth]{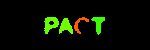}
    \caption{}
    \label{fig:mser_tree_tree_accumulation}
  \end{subfigure}

  \caption{MSERs pruning.
    (a) MSER tree of a text segment;
    (b) MSERs colored according to variations, as
    variations increase, MSERs are colored from green to
    yellow then to red;
    (c) MSERs colored according to regularized
    variations;
    (d) MSER tree after linear reduction;
    (e) character candidates after tree accumulation.
  }\label{fig:mser_reduction}
\end{figure}

Figure~\ref{fig:mser_tree_var} shows a MSER tree colored
according to variation.
As variation increases, the color changes from green to
yellow then to red.
The same tree colored according to regularized variation is
shown in Figure~\ref{fig:mser_tree_regularized_var}.
The MSER tree in Figure~\ref{fig:mser_tree_regularized_var}
are used in our linear reduction (result presented in Figure~\ref{fig:mser_tree_linear_reduction}) and tree accumulation
algorithm (result presented in Figure~\ref{fig:mser_tree_tree_accumulation}).
Notice that ``variation'' in the following sections refer to
``regularized variation''.

\subsubsection{Linear Reduction}

The linear reduction algorithm is used in situations where
MSERs has only one child.
The algorithm chooses from parent and child the one with the
minimum variation and discards the other.

This procedure is applied across the whole tree recursively.
The detailed algorithm is presented in
Figure~\ref{fig:linear_reduction}.
Given a MSER tree, the procedure
returns the root of the processed
tree whose linear segments are reduced.
The procedure works as follows.
Given a node $t$, the procedure checks the number of
children of $t$; if $t$ has no children, returns $t$ immediately;
if $t$ has only one child,
get the root $c$ of child tree by first applying
the linear reduction procedure to the child tree; if $t$ has a
lower variation compared with $c$, link $c$'s children to $t$
and return $t$; otherwise we return $c$; if $t$ has more than
one children, process these children using linear reduction
and link the resulting trees to $t$ before returning $t$.
Figure~\ref{fig:mser_tree_linear_reduction} shows the
resulting MSER tree
after applying linear reduction to the tree shown in
Figure~\ref{fig:mser_tree_regularized_var}.
Note that in the resulting tree
all linear segments are reduced and non-leaf nodes always
have more than one children.

\begin{figure}[htb!]
  \begin{algorithmic}[1]
    \Procedure{Linear-Reduction}{$T$}
    \If{nchildren[$T$] = 0}
    \State \textbf{return} $T$
    \ElsIf{nchildren[$T$] = 1}
    \State $c$ $\gets$ {\Call{Linear-Reduction}{child[$T$]}}
    \If{var[$T$] $\leq$ var[$c$]}
    \State{link-children($T$, children[$c$])}
    \State \textbf{return} $T$
    \Else
    \State \textbf{return} $c$
    \EndIf
    \Else\Comment{nchildren[$T$] $\geq$ 2}
    \For{ \textbf{each}  $c$  $\in$ children[$T$]}
    \State link-children($T$, {\Call{Linear-Reduction}{$c$}})
    \EndFor
    \State \textbf{return} $T$
    \EndIf
    \EndProcedure
  \end{algorithmic}
  \caption{The linear reduction algorithm.}
  \label{fig:linear_reduction}
\end{figure}

\subsubsection{Tree Accumulation}
The tree accumulation algorithm is used when MSERs has
more than one child.
Given a MSER tree, the procedure returns a set of
disconnected nodes.
The algorithm works as follows.
For a given node $t$, tree accumulation checks the number of
$t$'s children; if $t$ has no children, return $t$ immediately; if $t$
has more than two children, create an empty set $C$ and
append the result of applying tree accumulation to
$t$'s children to $C$;
if one of the nodes in $C$ has a lower variation
than  $t$'s variation, return $C$, else
discard $t$'s children and return $t$.
Figure~\ref{fig:mser_tree_tree_accumulation} shows the result of
applying tree accumulation to the tree shown in
Figure~\ref{fig:mser_tree_linear_reduction}.
Note that the final result is a set of disconnected nodes
containing all the characters in the original MSER tree.

\begin{figure}[htb!]
  \begin{algorithmic}[1]
    \Procedure{Tree-Accumulation}{$T$}
    \If{nchildren[$T$] $\geq$ 2}
    \State $C \gets \emptyset$
    \For{ \textbf{each} $c$ $\in$ children[$T$]}
    \State $C \gets C \ \cup $ {\Call{Tree-Accumulation}{$c$}}
    \EndFor
    \If{var[$T$] $\leq$ min-var[$C$]}
    \State discard-children($T$)
    \State \textbf{return} $T$
    \Else
    \State \textbf{return} $C$
    \EndIf
    \Else\Comment{nchildren[$T$] = 0}
    \State \textbf{return} $T$
    \EndIf
    \EndProcedure
  \end{algorithmic}
  \caption{The tree accumulation algorithm.}
  \label{fig:tree-accumulation}
\end{figure}

\subsubsection{Complexity Analysis}
\label{section:complexity_analysis}
The linear reduction and tree accumulation algorithm
effectively visit each nodes in the MSRE tree and do simple
comparisons and pointer manipulations, thus the complexity
is linear to the number of tree nodes.
The computational complexity of the variation regularization
is mostly due to the calculations of MSERs' bounding rectangles,
which is up-bounded by the number of pixels in the image.




\subsection{Text Candidates Construction}
\label{section:region_construction}

\subsubsection{Text Candidates Construction Algorithm
  Overview}
Text candidates are constructed by clustering character
candidates using the single-link
clustering algorithm~\cite{clustering}.
Intuitively, single-link clustering produce clusters that
are elongated~\cite{clustering_review} and thus is
particularly suitable for the text candidates construction task.
Single-link clustering belongs to the family of hierarchical
clustering; in hierarchical clustering,
each data point is initially treated as a
singleton cluster and clusters are successively merged
until all points have been merged into a single remaining
cluster.
In the case of single-link clustering, the two clusters
whose two closest
members have the smallest distance are merged in each step.
A distance threshold can be specified such that the clustering
progress is terminated when the distance between nearest
clusters exceeds the threshold.
The resulting clusters of single-link algorithm form
a hierarchical cluster tree or cluster forest if termination
threshold is specified.
In the above algorithm, each data point represent a
character candidate and \emph{top level} clusters in the
final hierarchical cluster tree (forest) correspond to
text candidates.

The problem is of course to determine the distance
function and threshold for the single-link algorithm.
We use the weighted sum of features as the distance
function.
Given two data points $u,v$, let $x_{u,v}$ be the feature
vector characterizing the similarity between $u$ and $v$, the distance
between $u$ and $v$ is defined as
\begin{equation}
  d(u,v;w) = w^T x_{u,v},
  \label{eq:metric00}
\end{equation}
where $w$, the feature weight vector together with the
distance threshold, can be learned using the proposed distance metric learning
algorithm.

In the following subsections, we first introduce the feature
space $x_{u,v}$, then detail the proposed metric learning algorithm
and finally present the empirical analysis on the proposed algorithm.

\subsubsection{Feature Space}
The feature vector $x_{u,v}$ is used to describe
the similarities between data points $u$ and $v$.
Let $x_u, y_u$ be the coordinates of top left corner of $u$'s bounding rectangle,
$h_u, w_u$ be the height and width of the bounding
rectangle of $u$,
$s_u$ be the stroke width of $u$,
$c1_u, c2_u, c3_u$ be the average three channel color value of $u$,
feature vector ${x}_{u,v}$ include the
following features:

\begin{itemize}
\item Spatial distance
  \begin{align*}
    abs(x_u + 0.5 h_u - x_v - 0.5 w_u) / \max(w_u , w_v).
  \end{align*}

\item Width and height differences
  \begin{align*}
    & abs(w_u - w_v) / \max(w_u , w_v), \\
    & abs(h_u - h_v) / \max(h_u , h_v).
  \end{align*}

\item Top and bottom alignments
  \begin{align*}
    & \arctan(\frac{abs(y_u - y_v)} { abs(x_u + 0.5 h_u - x_v - 0.5 w_u)}), \\
    & \arctan(\frac{abs(y_u  + h_u - y_v - h_v)} { abs(x_u + 0.5 h_u - x_v - 0.5 w_u)}).
  \end{align*}

\item Color difference
  \begin{align*}
    \sqrt{(c1_u - c1_v)^2 + (c2_u -
      c2_v)^2 + (c3_u - c3_v)^2}.
  \end{align*}

\item Stroke width difference
  \begin{align*}
    abs(s_u - s_v)
    / \max(s_u, s_v).
  \end{align*}
\end{itemize}


\subsubsection{Distance Metric Learning}

There are a variety of distance metric learning methods~\cite{huang1, huang2, huang3}.
More specifically, many clustering algorithms rely on a good distance metric
over the input space.
One task of semi-supervised clustering is to learn a
distance metric that satisfies the labels or constrains in the
supervised data given the clustering
algorithm~\cite{integrating,xing_metric,klein}.
The strategy of metric learning  is to the learn distance
function by minimizing
distance between point pairs in $\mathcal{C}$ while maximizing
distance between point pairs in $\mathcal{M}$,
where $\mathcal{C}$ specifies pairs of points in different
clusters and $\mathcal{M}$ specifies pairs of points in
the same cluster.
In  single-link clustering, because clusters are formed by merging
smaller clusters, the final resulting clusters will form a
binary hierarchical cluster tree, in which non-singleton
clusters have exactly two direct subclusters.
It is not hard to see that the following
property holds for \emph{top level} clusters: given the termination threshold $\epsilon$,
it follows that distances between each top level cluster' subclusters
are less or equal to $\epsilon$
and distances between data pairs in different top level clusters are
great than $\epsilon$, in which the distance between clusters
is that of the two closest members in each cluster.
This property of single-link clustering enables us to design
a learning algorithm that can learn the distance function and
threshold simultaneously.

Given the top level cluster
set $\{C_k\}_{k=1}^{m}$, we randomly initialize feature weights
$w$ and set $\mathcal{C}$ and $\mathcal{M}$ as
\begin{align}
  & \mathcal{C} = \{(\hat{u}_k,
  \hat{v}_k)= \argmin_{u \in C_k, v \in C_{-k}}d(u,v;w)\}_{k=1}^{m},
  \label{eq:cannot-link} \\
  & \mathcal{M} = \{(u_k^*, v_k^*) =
  \argmin_{u \in C_k^1, v \in C_k^2} d(u,v;w)\}_{k=1}^{m},
  \label{eq:must-link}
\end{align}
where $C_{-k}$ is the set of points excluding points in
$C_k$, $C_k^1$ and $C_k^2$ are direct subclusters of $C_k$.
Suppose $\epsilon$ is specified as the single-link
clustering termination threshold. By the definition of
single-link clustering, we must have
\begin{align}
  & d(u,v;w) > \epsilon \text{ for all } (u,v) \in
  \mathcal{C},
  \label{eq:cannot-link-constain}\\
  & d(u,v;w) \leq \epsilon \text{ for all } (u,v) \in
  \mathcal{M}.
  \label{eq:must-link-constain}
\end{align}
The above equations show that $\mathcal{C}$ and
$\mathcal{M}$ can be corresponded as the positive and
negative sample set of a classification problem, such that
feature weights and threshold can be learned by minimizing
the classification error.
As we know, the logistic regression loss is the traditional loss used in classification
with a high and stable performance.
By adopting the objective function of logistic regression,
we define the following objective function

\begin{align}
 \label{eq:logistic_regression1}
  J(\theta: \mathcal{C}, \mathcal{M}) = & \frac{-1}{2m} ( \sum_{(u,v) \in \mathcal{C}} \log(h_{\theta}(x'_{u,v})) \\ \notag
   & + \sum_{(u,v) \in \mathcal{M}} \log(1-h_{\theta}(x'_{u,v})) ),
\end{align}
where
\begin{align}
  & h_{\theta}(x'_{u,v}) = 1/(1+\exp(-\theta^T x'_{u,v})), \\ \nonumber
  & \theta = \left(
    \begin{array}{c}
      -\epsilon \\ \nonumber
      w
    \end{array}
  \right),\\ \nonumber
  & x'_{u,v} =
  \left(
    \begin{array}{c}
      1 \\ \nonumber
      x_{u,v}
    \end{array}
  \right).
\end{align}
The feature weights $w$ and
threshold $\epsilon$ can be learned simultaneously
by minimizing the objective function $J(\theta: \mathcal{M},
\mathcal{C})$ with respect to current assignment of $\mathcal{C}$ and $\mathcal{M}$
\begin{equation}
\label{eq:logistic_regression2}
\theta^* = \argmin_{\theta} J(\theta: \mathcal{C}, \mathcal{M})
\end{equation}
Minimization of the above objective function is a
typical nonlinear optimization problem and can be solved by
classic gradient optimization methods~\cite{elements_sl_book}.

Note that in the above learning scheme,
initial values for $w$ have to be
specified in order to generate set $\mathcal{C}$ and
$\mathcal{M}$ according to Equation~\eqref{eq:cannot-link}
and~\eqref{eq:must-link}.
For this reason, we design an iterative optimization algorithm in
which each iteration involves two successive steps
corresponding to assignments of $\mathcal{C}, \mathcal{M}$
and optimization with respect to $\mathcal{C}, \mathcal{M}$.
We call our algorithm as ``\emph{self-training distance metric learning}''.
Pseudocode for this learning algorithm is presented in
Figure~\ref{fig:metric-learning}.
Given the top level cluster set $\{C_k\}_{k=1}^{m}$, the
learning algorithm find an optimized $\theta$
such that the objective function $J(\theta:\mathcal{C},
\mathcal{M})$ is minimized with respect to $\mathcal{C}, \mathcal{M}$.
In this algorithm, initial value for $\theta$ is set before
the iteration begins; in the first stage of the iteration $\mathcal{M}$ and
$\mathcal{C}$ are update according to Equation~\eqref{eq:cannot-link}
and~\eqref{eq:must-link} with respect to current assignment
of $\theta$; in the second stage, $\theta$ is updated by
minimizing the objective function with respect to the
current assignment of $\mathcal{C}$ and $\mathcal{M}$.
This two-stage optimization is then repeated until convergence
or the maximum number of iterations is exceeded.
\begin{figure}[htb!]
  \begin{algorithmic}
    \State{\textbf{Input:} labeled clusters set $\{C_k\}_{k=1}^{m}$}
    \State{\textbf{Output:} optimized $\theta$ such that
      objective function $J$ is minimized}
    \State{\textbf{Method:}}
    \State{randomly initialize $\theta$}
    \Repeat
    \State{\textbf{stage1}: update $\mathcal{M}$ and
      $\mathcal{C}$ with respect to $\theta$ }
    \State{\textbf{stage2}: $\theta \gets
      \argmin_{\theta}J(\theta: \mathcal{C}, \mathcal{M})$}
    \Until{convergence or reach iteration limitation}
  \end{algorithmic}
  \caption{The self-training distance metric learning algorithm.}
  \label{fig:metric-learning}
\end{figure}

Similar to most self-training algorithms, convergence of the proposed algorithm is not guaranteed
because the objective function is not assured to decrease in
stage one. However, self-training algorithms have demonstrated their success in many applications. In our case, we find that our algorithm can usually generate very good performance
after a very small number of iterations, typically in 5 iterations. This phenomenon will be investigated in the next subsection.

\subsubsection{Empirical Analysis}
We perform an empirical analysis on the proposed  distance metric learning algorithm.
We labeled in the ICDAR 2011 competition dataset $466$ text candidates corresponding to true
text in the  training set, 70\% of which used as
training data, 30\% as validation data.
In each iteration of the algorithm,
cannot-link set $\mathcal{C}$ and must-link set $\mathcal{M}$ are updated
in step one by generating cannot-link point pairs and must-link
point pairs from true text candidates
in every image in the training dataset;
the objective function are optimized using the L-BFGS
method~\cite{LBFGS} and the parameters are updated in stage two.
Performance of the learned distance weights and threshold in
step two is evaluated on the validation dataset in each
iteration.

As discussed in the previous section, the algorithm may
or may not converge due to different initial values
of the parameters.
Our experiments show that the learned parameters almost always
have a very low error rate on the validation set after the first several iterations and no major
improvement is observed in the continuing iterations.
As a result, whether the algorithm converge or not has no great
impact on the performance of the learned parameters.

\begin{figure}[htb!]
  \centering
  \begin{subfigure}[b]{0.24\textwidth}
    \centering
    \includegraphics[width=\textwidth]{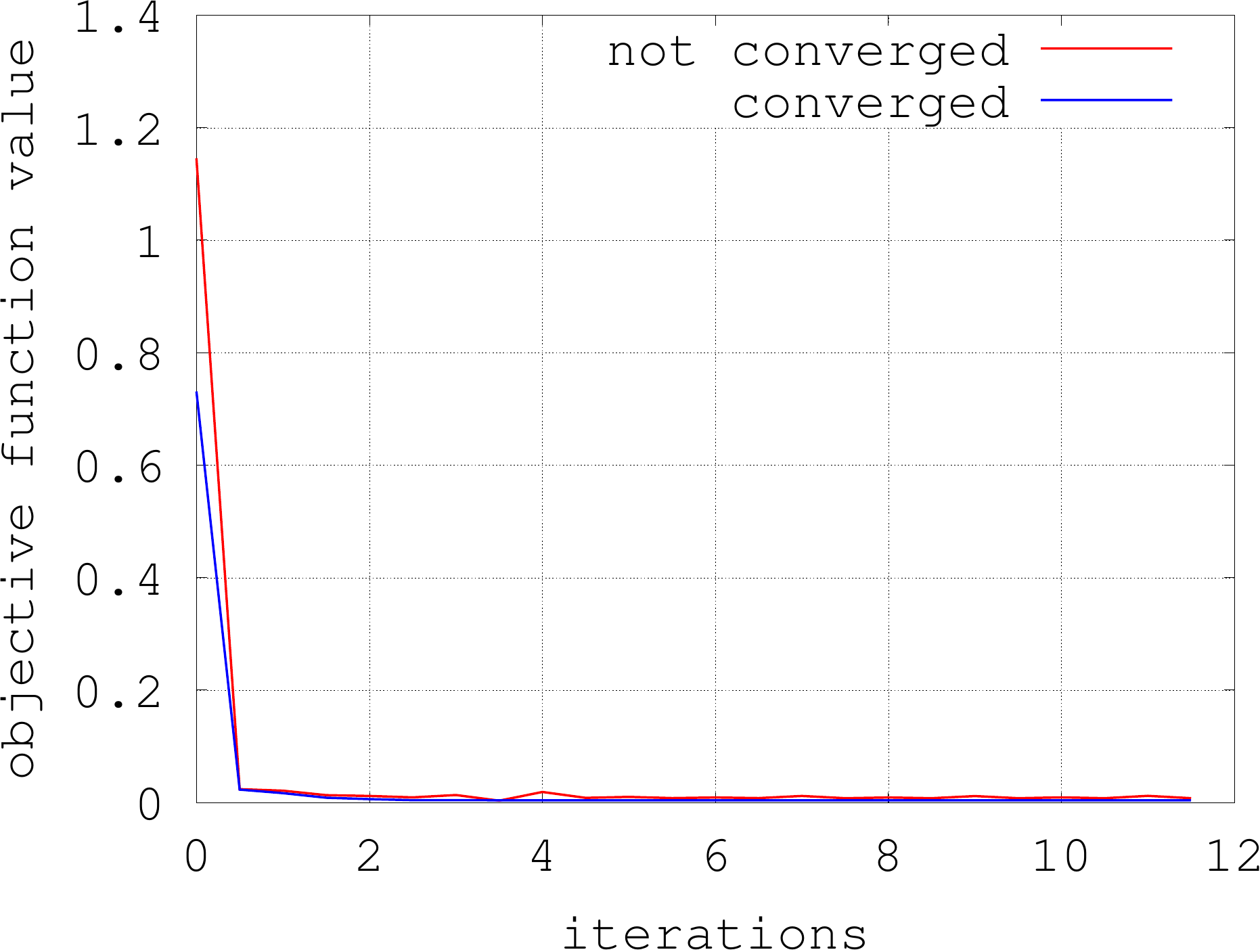}
    \caption{}
    \label{fig:objective}
  \end{subfigure}
  \begin{subfigure}[b]{0.24\textwidth}
    \centering
    \includegraphics[width=\textwidth]{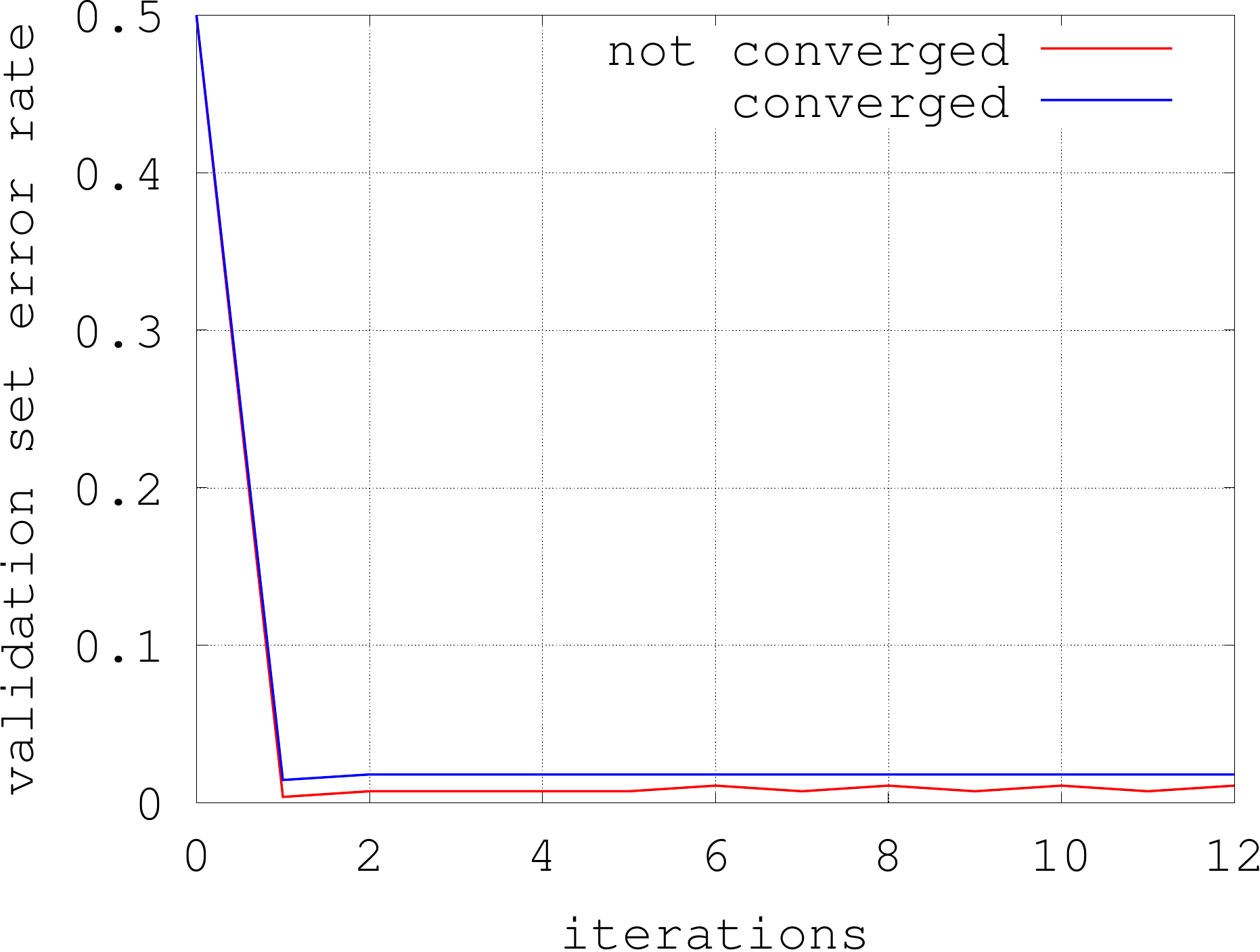}
    \caption{}
    \label{fig:error-rate}
  \end{subfigure}
  \caption{
    Objective function value (a) and validation set error
    rate of learned parameters (b)
    after stage one and stage two in each iteration of two instance  of the metric
    learning algorithm; the red line corresponds to the
    not converged instance and the blue line corresponds to
    the converged instance.
  }
  \label{situations}
\end{figure}

We plot the value of the objective
function after stage one and stage two in each iteration of
two instance (correspond to a converged one and not
converged one) of the
algorithm in Figure~\ref{fig:objective}.
The corresponding error rates of the learned parameters on
the validation set in each iteration are plotted in
Figure~\ref{fig:error-rate}.
Notice that  value of the objective function and
validation set error rate dropped immediately after the first several iterations.
Figure~\ref{fig:error-rate} shows that the learned
parameters have different error rates due to different
initial value, which suggests to run
the algorithm several times to get the satisfactory parameters.
The parameters for the single-link clustering algorithm in our scene text detection system are chosen
based on the performance on the validation set.

\subsection{Text Candidates Elimination}
\label{section:character_classifier}
Using the text candidates construction algorithm proposed in
Section~\ref{section:region_construction},
our experiment in ICDAR 2011 competition training set shows
that only 9\% 
of the text candidates correspond to true
text.
As it is hard to train an effective text classifier using such
unbalanced dataset, most of the non-text candidates need to
be removed before training the classifier.
We propose to use a character classifier to estimate the
posterior probabilities of text candidates corresponding to
non-text and remove text candidates with high probabilities for non-text.

The following features are used to train the character classifier.
Smoothness,  defined as the average difference of adjacent
boundary pixels' gradient directions, stroke width features, including the average stroke width and stroke width variation,
height, width, and aspect ratio.
Characters with small aspect ratios such as ``i'', ``j''
and ``l'' are treated as
negative samples, as it is very uncommon that some words
comprise many small aspect ratio characters.

Given a text candidate $T$, let $O(m,n;p)$ be the observation
that there are $m$ ($m \in \mathbb{N}, m \geq 2$) character
candidates in $T$, of which $n$ ($n \in \mathbb{N}, n \leq m$)
are classified as non-characters by a character classifier of
precision $p$ ($0 < p < 1$).
The probability of the observation conditioning on $T$
corresponding to text and non-text are $P(O(m,n;p) |
\text{text}) = p^{m - n} (1 - p)^{n}$ and $P(O(m,n;p) |
\text{non-text}) = (1 - p)^{m - n} p^{n}$ respectively.
Let $P(\text{text})$ and $P(\text{non-text})$ be the prior
probability of $T$ corresponding to text and non-text.
By applying Bayes' rule, the posterior probability of
$T$ corresponding to non-text given the observation is
\begin{align}
  P(\text{non-text} | &O(m,n;p)) = \nonumber \\
  & \frac{P(O(m,n;p) | \text{non-text}) P(\text{non-text})}{P(O(m,n;p))},
\end{align}
where $P(O(m,n;p))$ is the probability of the observation
\begin{align}
  P(O(m,n;p)) &= P(O(m,n;p) | \text{text}) P(\text{text})
  \nonumber \\
  &\qquad {} +  P(O(m,n;p) | \text{non-text})
  P(\text{non-text}),
\end{align}
The candidate region is rejected if
\begin{equation}
  P(\text{non-text} |O(m,n;p)) \geq \varepsilon,
\end{equation}
where $\varepsilon$ is the threshold.

Our experiment shows that text candidates of different sizes
tend to have different probability of being text.
For example, on the ICDAR training set, 1.25\% of text candiates
of size two correspond to text, while 30.67\% of text
candidates of size seven correspond to text, which suggests
to set different priors for text candidates of different
size.
Given a text candidates $T$ of size $s$, let $N_s$ be the total
number of text candidates of size $s$, $N_s^*$ be the number
of text candidates of size $s$ that correspond to text,
we estimate the prior of $T$ being text as $P_s(\text{text}) =
{N_s^*} / {N_s}$, and the prior of $T$ being non-text as
$P_s(\text{non-text}) = 1 - P_s(\text{text})$.
These priors are computed based on statistics on the ICDAR
training dataset.

\begin{figure}[htb!]
  \centering
  \begin{subfigure}[b]{0.24\textwidth}
    \centering
    \includegraphics[width=\textwidth]{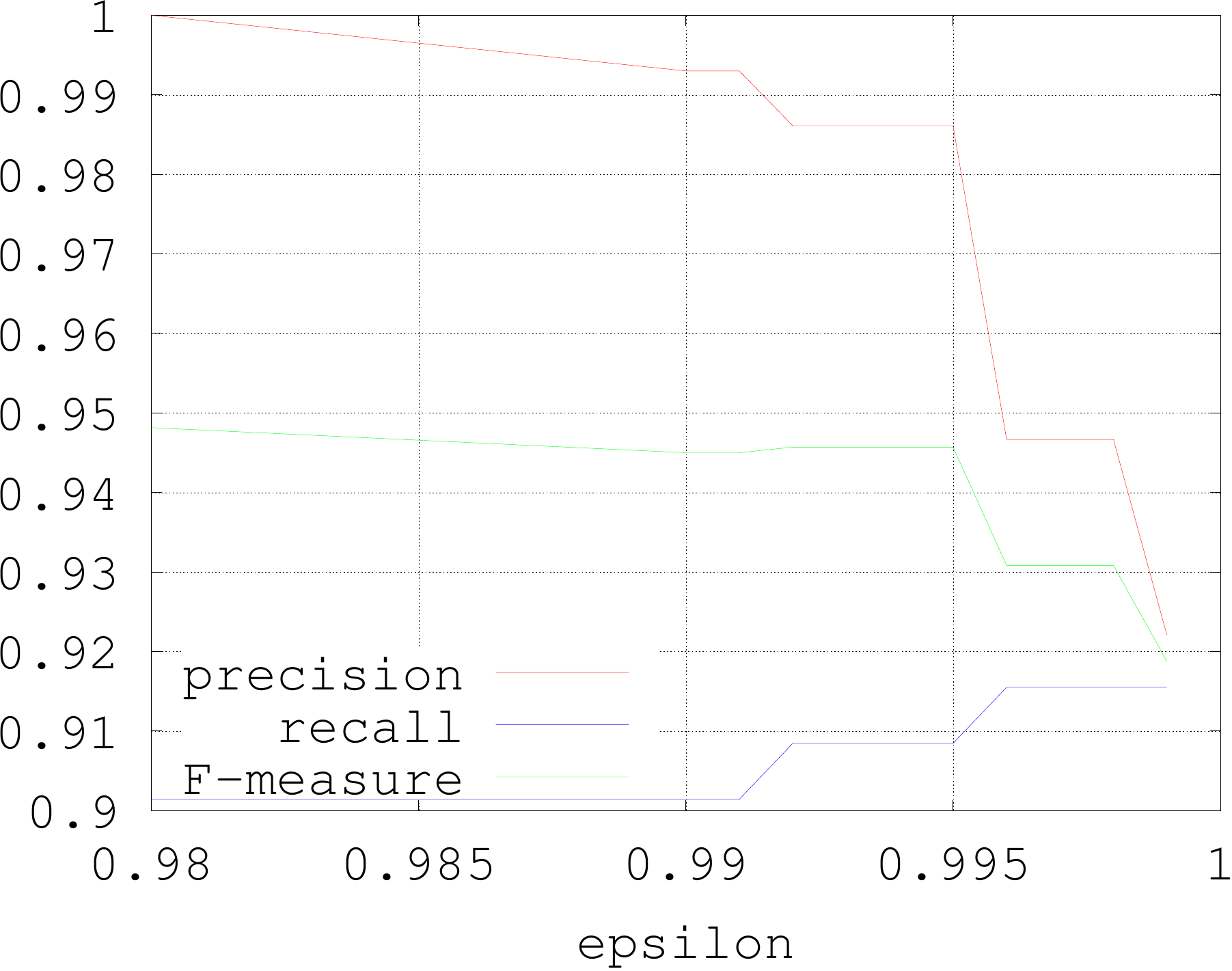}
    \caption{}
    \label{fig:precision_recall}
  \end{subfigure}
  \begin{subfigure}[b]{0.24\textwidth}
    \centering
    \includegraphics[width=\textwidth]{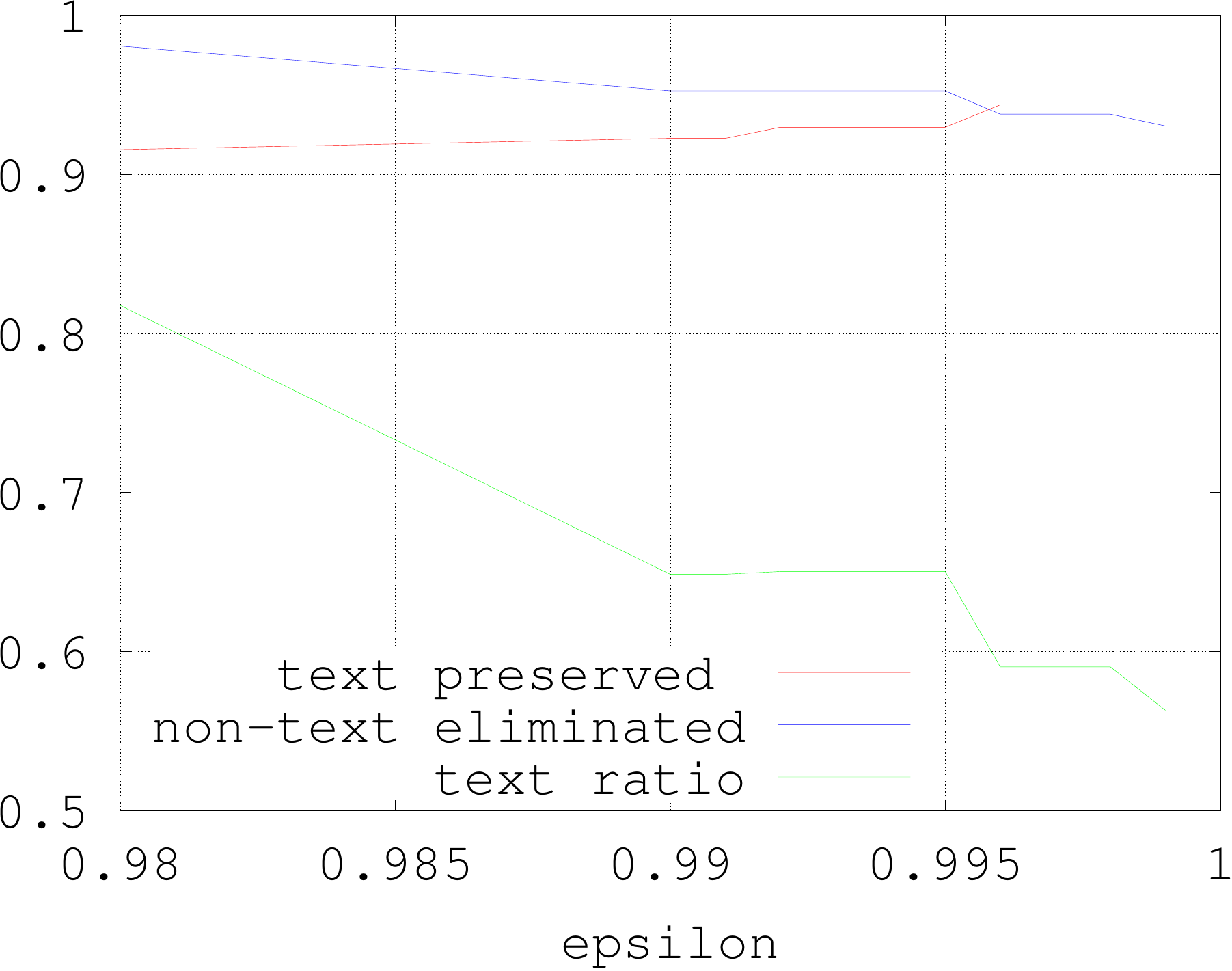}
    \caption{}
    \label{fig:elimination}
  \end{subfigure}
  \caption{Performance of different $\varepsilon$ on the
    validation set.
    (a) Precision, recall and $f$ measure of text
    classification task;
    (b) ratio of preserved text samples, ratio of eliminated
    non-text samples and ratio of text samples.
  }\label{fig:varepsilon_perfomance}
\end{figure}

To find the appreciate $\varepsilon$, we used 70\% of ICDAR
training dataset to train the character classifier and text classifier,
the remaining 30\% as validation set to test the performance
of different $\varepsilon$.
Figure~\ref{fig:precision_recall} shows the precision,
recall and $f$ measure of text candidates classification
task on the validation set.
As $\varepsilon$ increases, text candidates are more
unlikely to be eliminated, which results in the increase of
recall value. In the scene text detection task, recall is
preferred over precision, until $\varepsilon = 0.995$ is
reached, where a major decrease of $f$ measure occurred,
which can be explained by the sudden decrease of ratio of
text samples (see Figure~\ref{fig:elimination}).
Figure~\ref{fig:elimination} shows that at $\varepsilon =
0.995$, $92.95\%$ of text are preserved, while $95.25\%$ of
non-text are eliminated.

\section{Experimental Results}
\label{section:experimental_results}

\footnote{
  An online demo of the proposed scene text detection system
  is available at \emph{\url{http://kems.ustb.edu.cn/learning/yin/dtext}.}
}

In this section, we presented the experimental results of
the proposed scene text detection method on two publicly
available benchmark datasets, ICDAR 2011 Robust
Reading Competition dataset
\footnote{The ICDAR 2011 Robust
Reading Competition dataset is available at
\emph{\url{http://robustreading.opendfki.de/wiki/SceneText}}.}
and the multilingual dataset
\footnote{The multilingual dataset is available at \emph{\url{http://liama.ia.ac.cn/wiki/projects:pal:member:yfpan}}.}
provided by Pan et al.~\cite{pan}.

\subsection{Experiments on ICDAR 2011 Competition Dataset }
\label{section:icdar2011}

The ICDAR 2011 Robust Reading Competition (Challenge 2: Reading Text in
Scene Images) dataset ~\cite{icdar2011} is a widely used
dataset for benchmarking scene text detection algorithms.
The dataset contains $229$ training images
and $255$ testing images.
The proposed system is trained on the
training set and evaluated on the testing set.

It is worth noting that the evaluation scheme of ICDAR 2011
competition is not the same as of ICDAR 2003 and ICDAR
2005.
The new scheme, the \emph{object count/area} scheme
proposed by Wolf et al.~\cite{Wolf_Jolion_2006}, is more
complicated but offers
several enhancements over the old scheme.
Basically, these two scheme use the notation of precision,
recall and $f$ measure that is defined as
\begin{align}
  & recall = \frac{\sum_{i=1}^{|G|} match_G(G_i)}{|G|}, \\
  & precision = \frac{\sum_{j=1}^{|D|} match_D(D_j)}{|D|}, \\
  & f = 2 \frac{recall \cdot precision}{recall + precision},
\end{align}
where $G$ is the set of groundtruth rectangles and $D$ is
the set of detected rectangles.
In the old evaluation scheme, the matching functions are defined
as
\begin{align}
  & match_G(G_i) = \max_{j = 1...|D|} \frac{2 \cdot area(G_i
    \cap D_j)}{area(G_i) + area(D_j)}, \\
  & match_D(D_j) = \max_{i = 1...|G|} \frac{2 \cdot area(D_j
    \cap G_i)}{area(D_j) + area(G_i)}.
\end{align}
The above matching functions only consider one-to-one matches
between groundtruth and detected rectangles, leaving room
for ambiguity between detection quantity and
quality~\cite{Wolf_Jolion_2006}.
In the new evaluation scheme, the
matching functions are redesigned considering detection quality
and different matching situations (one-to-one matchings,
one-to-many matchings and many-to-one matchings) between groundtruth
rectangles and detected rectangles, such that the detection
quantity and quality can both be observed using the new
evaluation scheme.
The evaluation software DetEval \footnote
{DetEval is available at \emph{\url{http://liris.cnrs.fr/christian.wolf/software/deteval/index.html}}.}
used by ICDAR 2011 competition is available online and free to use.

\begin{figure}[htb!]
  \centering
  \includegraphics[width=0.5\textwidth]{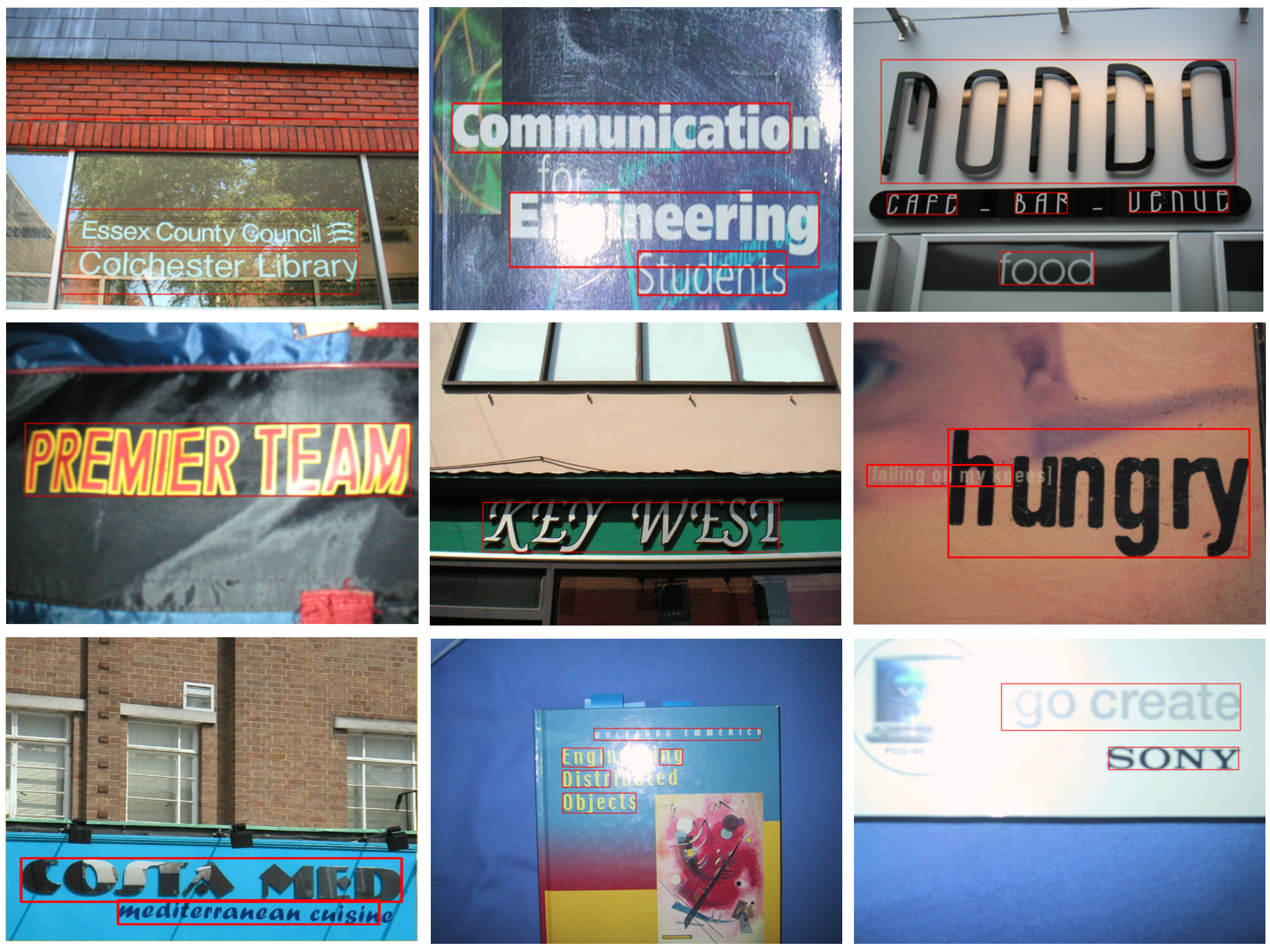}
  \caption{Text detection examples on the ICDAR 2011
    dataset.
    Detected text by our system are labeled using red rectangles.
    Notice the robustness against low contrast, complex
    background and font variations.}
  \label{fig:samples_icdar}
\end{figure}

The performance of our system, together with Neumann and Matas'
method~\cite{real_time}, a very recent MSER based method by
Shi et al.~\cite{mser2013}
and some of the top
scoring methods
(Kim's method, Yi's method, TH-TextLoc system and Neumann's method)
from ICDAR 2011 Competition
are presented in
Table~\ref{table:performance}.
As can be seen from Table~\ref{table:performance},
our method produced much better recall, precision and
$f$ measure over other methods on this dataset.
It is worth noting that the first four methods in
Table~\ref{table:performance} are all MSER based methods and
Kim's method is the winning method of ICDAR 2011 Robust
Reading Competition.
Apart from the detection quality, the proposed system offers
speed advantage over some of the listed methods.
The average processing speed of the proposed system on a
Linux laptop with Intel (R) Core (TM)2 Duo 2.00GHZ CPU is 0.43s
per image.
The processing speed of Shi et al.'s method~\cite{mser2013}
on a PC with Intel (R) Core (TM)2 Duo 2.33GHZ CPU is 1.5s
per image.
The average processing speed of Neumann and Matas'
method~\cite{real_time} is 1.8s per image on a ``standard
PC''.
Figure~\ref{fig:samples_icdar} shows some text detection
examples by our system on ICDAR 2011 dataset.

\begin{table}[htb!]
  \centering
  \caption{Performance ($\%$) comparison of text detection algorithms on ICDAR 2011 Robust Reading Competition dataset.}
  \label{table:performance}
  \begin{tabular} {|c|c|c|c|}
    \hline
    Methods                            & Recall         & Precision      & $f$            \\ \hline
    \textbf{Our Method}               & \textbf{68.26} & \textbf{86.29} & \textbf{76.22} \\ \hline
    Shi et al.'s method~\cite{mser2013} & 63.1          & 83.3           & 71.8           \\ \hline
    Kim's Method (not published)      & 62.47          & 82.98          & 71.28          \\ \hline
    Neumann and Matas~\cite{real_time} & 64.7           & 73.1           & 68.7           \\ \hline
    Yi's Method                       & 58.09          & 67.22          & 62.32          \\ \hline
    TH-TextLoc System                 & 57.68          & 66.97          & 61.98          \\ \hline
    Neumann's Method                  & 52.54          & 68.93          & 59.63          \\ \hline
  \end{tabular}
\end{table}

To fully appreciate the benefits of
\emph{text candidates elimination}
and \emph{the MSERs pruning algorithm},
we further profiled the proposed system on this dataset using the following
schemes (see Table~\ref{table:component_profile})

1) \textbf{Scheme-I}, no text candidates elimination
performed. As can be seen from
Table~\ref{table:component_profile}, the absence of text
candidates elimination results in a major decrease in
precision value.
The degradation can be explained by the fact that
large number of non-text are
passed to the text candidates classification stage without
being eliminated.

2) \textbf{Scheme-II}, using default parameter setting~\cite{vlfeat} for the
MSER extraction algorithm.
The MSER extraction algorithm is controlled by several
parameters~\cite{vlfeat}: $\Delta$ controls how the variation is
calculated; maximal variation $v_{+}$ excludes too unstable MSERs;
minimal diversity $d_{+}$ removes duplicate MSERs by measuring the size difference
between a MSER and its parent.
As can be seen from Table~\ref{table:component_profile},
compared with our parameter setting ($\Delta = 1, v_+=0.5, d_+=0.1$),
the default parameter setting ($\Delta = 5, v_+=0.25,
d_+=0.2$) results in a major decrease in recall value.
The degradation can be explained by two reasons:
(1) the MSER algorithm is not able to detect some
low contrast characters (due to $v_+$), and
(2) the MSER algorithm tends to miss some regions that are
more likely to be characters (due to $\Delta$ and $d_+$).
Note that the speed loss (from 0.36 seconds to 0.43 seconds)
is mostly due to the MSER detection algorithm itself.





\begin{table}[htb!]
  \centering
  \caption{Performance (\%) of the proposed method due to
    different components}
  \label{table:component_profile}
  \begin{tabular} {|c|c|c|c|c|}
    \hline
    Component          & Recall & Precision & $f$   & Speed (s) \\ \hline
    Overall system     & 68.26  & 86.29     & 76.22 & 0.43      \\ \hline
    \textbf{Scheme-I}  & 65.57  & 77.49     & 71.03 & 0.41      \\ \hline
    \textbf{Scheme-II} & 61.63  & 85.78     & 71.72 & 0.36      \\ \hline
  \end{tabular}
\end{table}

\subsection{Experiments on Multilingual Dataset}
\begin{figure}[htb!]
  \centering
  \includegraphics[width=0.5\textwidth]{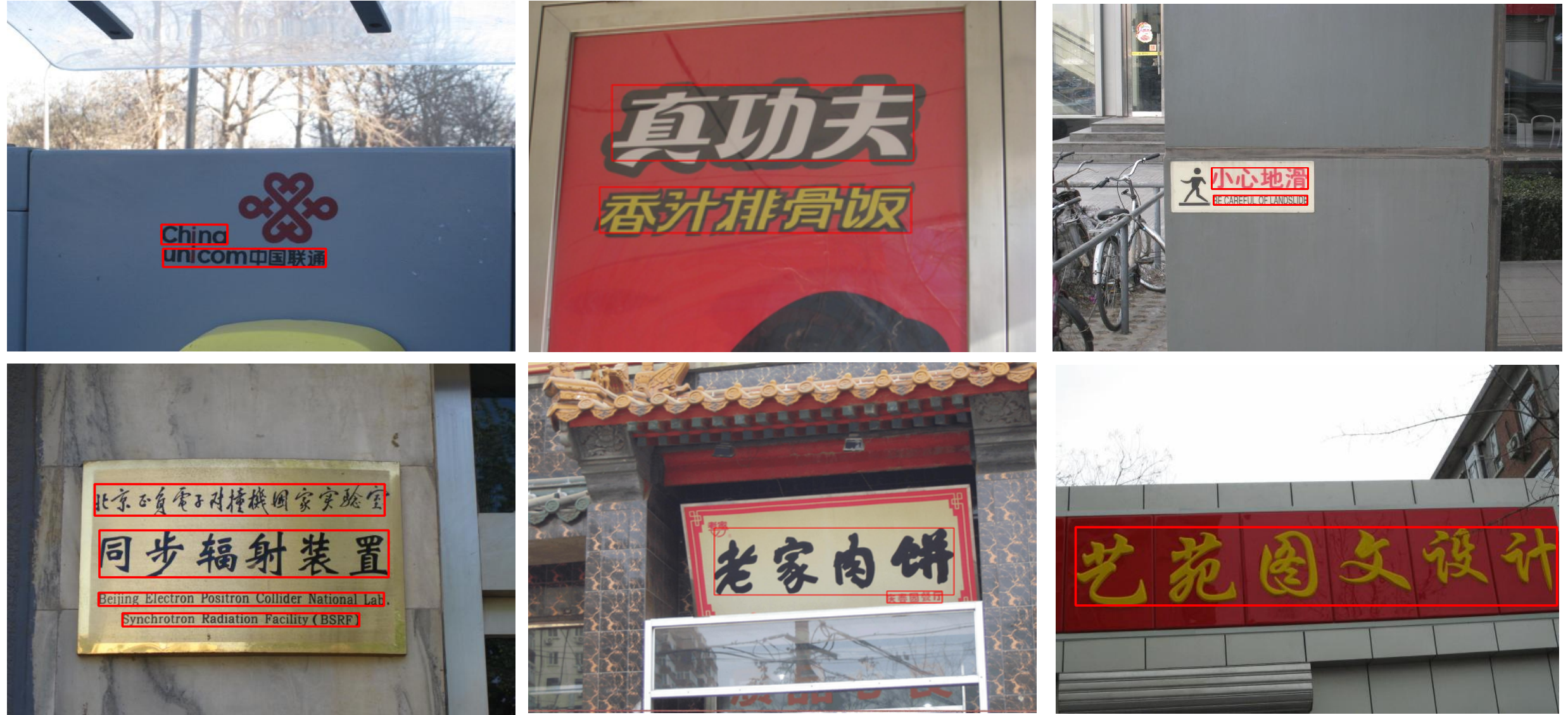}
  \caption{Text detection examples on the multilingual
    dataset. Detected text by our system are labeled using red rectangles.
  }
  \label{fig:samples_multilingual}
\end{figure}

The multilingual (include Chinese and English, see
Figure~\ref{fig:samples_multilingual}) dataset was
initially published by Pan et al.~\cite{pan} to evaluate the
performance of their scene text detection system.
The training dataset contains $248$ images and the testing
dateset contains $239$ images.
As there are no apparent spacing between Chinese word,
this multilingual dataset only provides groundtruths for
text lines.
We hence evaluate the text line detection performance of
our system without further partitioning text into words.
Figure~\ref{fig:samples_multilingual} shows some scene text
detection examples by our system on this dataset.

\begin{table}[htb!]
  \centering
  \caption{Performance ($\%$) comparison of text detection
    algorithms on the multilingual dataset.
    Speed of Pan et al.'s method is profiled on a
    PC with Pentium D 3.4GHz CPU.
  }
  \label{table:performance_multilingual}
  \begin{tabular} {|c|c|c|c|c|}
    \hline
    Methods                       & Recall  & Precision & $f$     & Speed (s)  \\ \hline
    \textbf{Scheme-III}                    & {63.23} & {79.38}   & {70.39} & {0.22} \\ \hline
    \textbf{Scheme-IV}                   & {68.45} & {82.63}   & {74.58} & {0.22} \\ \hline
    Pan et al.'s method~\cite{pan} & 65.9    & 64.5      & 65.2    & 3.11   \\ \hline
  \end{tabular}
\end{table}

The performance of our system (include \textbf{Scheme-III} and
\textbf{Scheme-IV}) and Pan et al.'s method~\cite{pan} is presented in
Table~\ref{table:performance_multilingual}.
The evaluation scheme in ICDAR 2003 competition (see
Section~\ref{section:icdar2011}) is used for fair comparison.
The main difference between \textbf{Scheme-III} and
\textbf{Scheme-IV} is that
the character classifier in the first scheme is trained on
the ICDAR 2011 training set while the character classifier in
the second scheme is trained on the multilingual training set
(character features for training the classifier are the same).
The result comparison between \textbf{Scheme-III} and
\textbf{Scheme-IV} in
Table~\ref{table:performance_multilingual}
shows that the performance of the proposed system is
significantly improved because of the incorporating of the
Chinese-friendly character classifier.
The basic implication of this improvement is that the
character classifier has a significant
impact on the performance of the overall system, which
offers another advantage of the proposed system: the
character classifier can be trained on desired dataset until
it is accurate enough and be plugged into the system and
the overall performance will be improved.
Table~\ref{table:performance_multilingual} also shows the
advantages of the proposed method over Pan et al.'s method
in detection quality and speed.


\section{Conclusion}
\label{section:conclusion}
This paper presents a new MSER based scene text
detection method.
Several key improvement over traditional methods have been
proposed.
We propose a fast and accurate MSERs pruning algorithm that
enables us to detect most the characters even when the image is
in low quality.
We propose a novel self-training distance metric learning algorithm that can
learn distance weights and threshold simultaneously;
text candidates are constructed by clustering character
candidates by the single-link algorithm using the learned
parameters.
We propose to use a character classifier to estimate the
posterior probability of text candidate corresponding to non-text and
eliminate text candidates with high probability for non-text, which helps
to build a more powerful text classifier.
By integrating the above ideas, we built a robust scene text
detection system that exhibited superior performance over
state-of-the-art methods on both the ICDAR 2011 Competition dataset and a multilingual dataset.


%


\ifCLASSOPTIONcompsoc
  \section*{Acknowledgments}
\else
  \section*{Acknowledgment}
\fi

The research was partly supported by National Basic Research Program of China (2012CB316301)
and National Natural Science Foundation of China (61105018, 61175020).

\ifCLASSOPTIONcaptionsoff
  \newpage
\fi



\bibliographystyle{IEEEtran}
\bibliography{IEEEabrv,./IEEEexample}
%

%




\end{document}